\documentclass[smallcondensed]{svjour3}                     
\smartqed  
\usepackage{graphicx}
\usepackage{url}
\usepackage{algorithm}
\usepackage{algorithmic}

\begin{document}

\title{An application of the Threshold Accepting metaheuristic for curriculum based course timetabling
}

\titlerunning{Threshold Accepting for curriculum based course timetabling}        

\author{Martin Josef Geiger}


\institute{M. J. Geiger \at
              University of Hohenheim\\
              70593 Stuttgart, Germany\\
              \email{mjgeiger@uni-hohenheim.de}
}

\date{Received: date / Accepted: date}

\maketitle

\begin{abstract}
The article presents a local search approach for the solution of
timetabling problems in general, with a particular implementation
for competition track 3 of the International Timetabling Competition
2007 (ITC 2007). The heuristic search procedure is based on
Threshold Accepting to overcome local optima. A stochastic
neighborhood is proposed and implemented, randomly removing and
reassigning events from the current solution.

The overall concept has been incrementally obtained from a series of
experiments, which we describe in each (sub)section of the paper. In
result, we successfully derived a potential candidate solution
approach for the finals of track 3 of the ITC 2007.

\end{abstract}

\section{Introduction}
{\em Timetabling} describes a variety of notoriously difficult
optimization problem with considerable practical impact. Important
areas within this context include employee timetabling, sport
timetabling, flight scheduling, and timetabling in universities and
other institutions of (often higher) education
\cite{carter:2001:incollection}.

Typically, timetabling is concerned with the assignment of
activities to resources. In more detail, these resources provide
timeslots (time intervals) to which the activities may be assigned
subject to certain side constraints. The overall objective of the
problem is to find a feasible assignment of all events such that
some desirable properties are present in the final solution.

Timetabling problems are challenging not only in terms of their
complexity, but also as they often involve multiple conflicting
objectives \cite{petrovic:2002:incollection} and even multiple
stakeholder with conflicting interests and views. University
timetabling problems present an interesting example of this problem
domain. Here, compromise solutions must be found that equally meet
the expectations of students and teachers.

Numerous publications are devoted to problem domain of timetabling,
with important work by the EURO Working Group on Automated
Timetabling WATT. Members of the group maintain a bibliography and
collect other timetabling-related resources under
\url{http://www.asap.cs.nott.ac.uk/watt/}.

More recently, timetabling competitions stimulated the scientific
development of the field, encouraging researchers to propose
solution approaches for newly released benchmark instances. By
creating a competitive atmosphere for algorithmic development,
similar to the famous DIMACS implementation challenges, fresh ideas
have been developed. In 2007, another timetabling competition
started, and this article describes a contribution and the obtained
results for it.

The article is organized as follows. In the following
Section~\ref{sec:problem}, the problem under investigation is
briefly described. An approach for the construction of initial
feasible solutions is presented in Section~\ref{sec:construction},
and experimental results of this constructive approach obtained on
benchmark instances are presented. The initially constructed
solutions are then improved using the iterative local search
heuristic given in Section~\ref{sec:iteration}. Experimental results
of the iterative phase are reported. Conclusions follow in
Section~\ref{sec:conclusions}.

\section{\label{sec:problem}The curriculum based timetabling problem}
The curriculum based timetabling problem
\cite{digaspero:2007:techreport} is a particular variant of an
educational timetabling problem, described in track 3 of the
International Timetabling Competition
(\url{http://www.cs.qub.ac.uk/itc2007/}).

It reflects the situation of many universities, where curricula
describe sets of courses such that any pair of courses of a
curriculum have students in common. Contrary to post-enrollment
based timetabling problems, where students register for courses they
wish to attend, some prior knowledge about the courses attended by
groups of students is required here. However, as university
faculties define the required courses that students have to attend,
this information is usually known.

A technical description of the problem is given in
\cite{digaspero:2007:techreport}. Besides some usual hard
constraints, four `soft constraints' are relevant that measure
desirable properties of the solutions, and it becomes clear that
these desirable properties of timetables are beneficial for both the
students as well as the lecturers:

\begin{enumerate}
\item A room capacity soft constraint tries to ensure that the number
of students attending a lecture does not exceed the room capacity.

\item Lectures must be spread into a minimum number of days,
penalizing timetables in which lectures appear in too few distinct
days.

\item The curricula should be compact, meaning that isolated
lectures, that is lectures without another adjacent lecture, should
be avoided.

\item All lectures of a course should be held in exactly one room.
\end{enumerate}

The overall evaluation of the timetables is then based on a weighted
sum approach, combining all four criteria in a single evaluation
function. While we adopt this approach in the current article, is
should be mentioned that Pareto-based approaches may be used as an
alternative way to handle the multi-criteria nature of the problem.

\section{\label{sec:construction}Construction of feasible initial solutions}
\subsection{Preprocessing}
Prior to the computation of a first solution, some preprocessing is
carried out. This preprocessing is relevant both for the
construction of an initial solution, as well as for the following
improvement phase. In brief, some problem-specific characteristics
are employed, adding some additional structure to the problem.

For each given lecture $L_{i}$, events $E_{i1}, \ldots, E_{ie}$ are
created which are later assigned to timeslots. The number of events
$e$ is given in the problem instances. Creating events for each
lecture leads to a more general problem description, and the
solution approach only needs to concentrate on the assignment of all
events, one to a single timeslot, as opposed to keeping track of
assigning a lecture to $e$ timeslots.

Second, we categorize for each lecture $L_{i}$ (and thus for each
event belonging to lecture $L_{i}$) the available rooms in three
disjunct classes $\mathcal{R}_{i1}, \mathcal{R}_{i2},
\mathcal{R}_{i3}$.\\$\mathcal{R}_{i1}$ refers to the rooms in which
the lecture fits best, that is the rooms $R_{k}$ with the minimum
positive or zero value of $c_{k} -s_{i}$,  $c_{k}$ being the  room
capacity, $s_{i}$ the number of students of lecture $L_{i}$.\\The
class $\mathcal{R}_{i2}$ stores the rooms in which lecture $L_{i}$
fits, that is $s_{i} < c_{k}$, but not best, and $\mathcal{R}_{i3}$
contains the rooms in which lecture $L_{i}$ does not fit. With
respect to the given problem statement, events of lectures may be
assigned to timeslots of rooms in $\mathcal{R}_{i3}$, this however
results in a penalty.

The underlying assumption of the classification of the rooms is that
events are preferably assigned to timeslots belonging to a room of
class $\mathcal{R}_{i1}$, followed by $\mathcal{R}_{i2}$ and
$\mathcal{R}_{i3}$. It has to be mentioned however, that this cannot
be understood as a binding, general rule but rather should be seen
as a recommendation. A randomized procedure is therefore going to be
implemented when assigning events to timeslots (see the following
section), allowing a certain deviation from the computed room order.

\subsection{A myopic construction approach}
\paragraph{The method}\hspace{0cm}\\%
The constructive phase tries to obtain a first feasible assignment
of all events to timeslots. A simple heuristic approach is used,
successively assigning all events to timeslots, one at a time, with
the given pseudo-code of Algorithm~\ref{alg:construction:1}. In this
description, we denote the set of all events with $\mathcal{E}$, and
the set of unassigned (open) events with $\mathcal{E}^{o}$. During
the successive assignment procedure, a set of events that have been
impossible to assign is maintained, denoted with $\mathcal{E}^{u}$.
In cases of assigning all events to timeslots, $\mathcal{E}^{u} =
\emptyset$ is returned.

\begin{algorithm}[!ht]
\caption{\label{alg:construction:1}Myopic construction}
\begin{algorithmic}[1]
\STATE{Set $\mathcal{E}^{o} = \mathcal{E}$}%
\STATE{$\mathcal{E}^{u} \leftarrow \emptyset$}
    \WHILE{$\mathcal{E}^{o} \neq \emptyset$}
        \STATE{Select the most critical event $E$ from $\mathcal{E}^{o}$, that is the event with the smallest number of available timeslots}
        \IF{$E$ can be assigned to at least one timeslot}
            \STATE{Select some available timeslot $T$ for $E$}
            \STATE{Assign $E$ to the timeslot $T$}
        \ELSE
            \STATE{$\mathcal{E}^{u} \leftarrow \mathcal{E}^{u} \cup E$}
        \ENDIF
        \STATE{$\mathcal{E}^{o} \leftarrow \mathcal{E}^{o} \backslash E$}
    \ENDWHILE
\end{algorithmic}
\end{algorithm}

A greedy approach is used in the assignment procedure, selecting in
each step the `most critical' event $E$ from $\mathcal{E}^{o}$, that
is the event with the smallest number of timeslots to which it may
be assigned.

The choice of timeslots for the events reflects the initial
categorization of rooms. With a probability of 0.5, timeslots of
rooms in $\mathcal{R}_{i1}$ are preferred over $\mathcal{R}_{i2}$
over $\mathcal{R}_{i3}$, and with a probability of 0.5, timeslots of
$\mathcal{R}_{i2}$ are preferred over the ones of $\mathcal{R}_{i1}$
over $\mathcal{R}_{i3}$. Within each class, timeslots are randomly
chosen with equal probability. In cases where a most-preferred class
of timeslots is empty, the choice is made from the lesser preferred
class and so on.\\As mentioned above, timeslots of rooms of class
$\mathcal{R}_{i1}$ are preferable to the ones of class
$\mathcal{R}_{i2}$ and $\mathcal{R}_{i3}$. The randomized assignment
procedure generally considers this aspect, however allowing a
certain deviation from the rule. This is done as we have been able
to observe that the assignment of events to timeslots following only
a single order does not lead to satisfactory results. In this case,
the choice of timeslots simply is too restrictive.\\It has been
pointed out in this context that the probability of assigning events
to timeslots of $\mathcal{R}_{i1} \rightarrow \mathcal{R}_{i2}
\rightarrow \mathcal{R}_{i3}$ could be expected to be greater than
the one of the order $\mathcal{R}_{i2} \rightarrow \mathcal{R}_{i1}
\rightarrow \mathcal{R}_{i3}$. While we generally agree with this
comment, other probabilities than 0.5 for both orders have not been
investigated yet. Consequently, subsequent experiments
certainly will have to examine the influence of this control
parameter on the obtained results.

\paragraph{Experimental results}\hspace{0cm}\\%
The constructive approach has been tested on the first seven
benchmark instances of ITC 2007 track 3. These are the instances
that initially have been made available by the organizers of the
competition. In February 2007, only a few weeks before the
submission deadline, seven more instances followed ({\tt
comp08.ctt}--{\tt comp14.ctt}). Obviously, experimental
investigations had to start considerable earlier, and we therefore
had to conclude on the effectiveness of the approach based on these
early seven instances.

After 1000 repetitions on each benchmark instance, we computed the
number of trials in which all events have successfully been assigned
to timeslots, given in Table~\ref{tbl:results:assignment:1}.

\begin{table}[!ht]
\caption{\label{tbl:results:assignment:1}Number of trials in which
all events have successfully been assigned (out of 1,000 trials)}
\begin{tabular}{lr}
\hline\noalign{\smallskip}%
Instance & Cases with $\mathcal{E}^{u} = \emptyset$\\%
\noalign{\smallskip}\hline\noalign{\smallskip}%
comp01.ctt & 1,000\\%
comp02.ctt &   354\\%
comp03.ctt &   377\\%
comp04.ctt & 1,000\\%
comp05.ctt &     0\\%
comp06.ctt &   953\\%
comp07.ctt &   827\\
\noalign{\smallskip}\hline%
\end{tabular}
\end{table}

The results reveal significant differences between the instances.
While we have been able to always assign all events to timeslots for
instance {\tt comp01.ctt} and {\tt comp04.ctt}, {\tt comp05.ctt}
turns out to be particularly difficult (constrained). After not
having been able to identify a single constructive run in which all
events have been assigned to timeslots, we conclude that simply
relying on more repetitions is most probably insufficient for this
instance. We rather need to adapt the constructive methodology to
the particular instance, overcoming problems with the assignment of
events to timeslots.

\subsection{Reactive repetitive reconstruction}
\paragraph{The method}\hspace{0cm}\\%
Based on the initial constructive approach, we propose a reactive
procedure that adapts to the set of unassigned events from previous
runs. The logic behind this approach is that the constructive
procedure `discovers' events that are difficult to assign, giving
them priority in successive runs. Similar ideas have been sketched
by the {\em squeaky wheel optimization} approach
\cite{joslin:1999:article}, and implemented in ant colony
metaheuristics for examination timetabling problems
\cite{dowsland:2005:article}.

In the following, let $\mathcal{E}^{p}$ be the set of prioritized
events, $\mathcal{E}^{\neg p}$ the set of non-prioritized events,
and $\mathcal{E}^{u}$ the set events that have not been assigned
during the construction phase. It is required that $\mathcal{E}^{p}
\subseteq \mathcal{E}$, $\mathcal{E}^{\neg p} \subseteq
\mathcal{E}$, $\mathcal{E}^{p} \cap \mathcal{E}^{\neg p} =
\emptyset$, and $\mathcal{E}^{p} \cup \mathcal{E}^{\neg p} =
\mathcal{E}$.

Algorithm~\ref{alg:construction:2} describes the reactive
construction procedure.

\begin{algorithm}[!ht]
\caption{\label{alg:construction:2}Reactive construction}
\begin{algorithmic}[1]
\STATE{Set $\mathcal{E}^{p} = \emptyset$, $\mathcal{E}^{u} =
\emptyset$, $loops = 0$}
\REPEAT%
    \STATE{$\mathcal{E}^{p} \leftarrow \mathcal{E}^{u}$}
    \STATE{$\mathcal{E}^{u} \leftarrow \emptyset$}
    \STATE{$\mathcal{E}^{\neg p} \leftarrow \mathcal{E} \backslash \mathcal{E}^{p}$}
    \WHILE{$\mathcal{E}^{p} \neq \emptyset$}
        \STATE{Select the most critical event $E$ from $\mathcal{E}^{p}$, that is the event with the smallest number of available timeslots}
        \IF{$E$ can be assigned to at least one timeslot}
            \STATE{Select some available timeslot $T$ for $E$}
            \STATE{Assign $E$ to the timeslot $T$}
        \ELSE
            \STATE{$\mathcal{E}^{u} \leftarrow \mathcal{E}^{u} \cup E$}
        \ENDIF
        \STATE{$\mathcal{E}^{p} \leftarrow \mathcal{E}^{p} \backslash E$}
    \ENDWHILE
    \WHILE{$\mathcal{E}^{\neg p} \neq \emptyset$}
        \STATE{Select the most critical event $E$ from $\mathcal{E}^{\neg p}$, that is the event with the smallest number of available timeslots}
        \IF{$E$ can be assigned to at least one timeslot}
            \STATE{Select some available timeslot $T$ for $E$}
            \STATE{Assign $E$ to the timeslot $T$}
        \ELSE
            \STATE{$\mathcal{E}^{u} \leftarrow \mathcal{E}^{u} \cup E$}
        \ENDIF
        \STATE{$\mathcal{E}^{\neg p} \leftarrow \mathcal{E}^{\neg p} \backslash E$}
    \ENDWHILE
    \STATE{$loops \leftarrow loops + 1$}
\UNTIL{$\mathcal{E}^{u} = \emptyset$ {\bf or} $loops = Maxloops$}
\end{algorithmic}
\end{algorithm}

As given in the pseudo-code, the construction of solutions is
carried out in a loop until either a feasible solution is identified
or a maximum number of iterations $Maxloops$ is reached. When
constructing a solution, a set of events $\mathcal{E}^{u}$ is kept
for which no timeslot has been found. When reconstructing a
solution, these events are prioritized over the others. In that
sense, the constructive approach is biased by its previous runs,
identifying events that turn out to be difficult to assign.

After at most a maximum number of $Maxloops$ iterations, the
construction procedure returns a solution that is either feasible
($\mathcal{E}^{u} = \emptyset$) or not ($\mathcal{E}^{u} \neq
\emptyset$).

It has been pointed out that even when events are put into $E^{p}$,
they do not necessarily remain elements of that set. Instead, they
might be removed from $E^{p}$ in the subsequent loop. To some
extent, this is counterintuitive, as the algorithm does not build up
a complete datastructure storing {\em all} unsuccessfully assigned
events. Instead, the direct `learning' is limited to the preceding
run. It has to be mentioned however, that some implicit information
is nevertheless transferred from loop to loop, as any loop is biased
by its predecessor. It also should be noticed that this
implementation of a more limited adaptive algorithm led to
satisfactory results, which is why alternative approaches have not
been further investigated yet.

\paragraph{Experimental results}\hspace{0cm}\\%
In the experiments, we focused on the difficult instance {\tt
comp05.ctt}, computing for 1000 trials the number of feasible
solutions reached after a certain number of loops of the
constructive approach. The obtained results are given in
Table~\ref{tbl:results:assignment:2}.

\begin{table}[!ht]
\caption{\label{tbl:results:assignment:2}Feasible solutions after a
certain number of loops for {\tt comp05.ctt} (out of 1,000 trials)}
\begin{tabular}{lr}
\hline\noalign{\smallskip}%
Loops & feasible solutions\\%
\noalign{\smallskip}\hline\noalign{\smallskip}%
1 &   0\\%
2 &  56\\%
3 & 272\\%
4 & 387\\%
5 & 511\\%
6 & 608\\%
7 & 688\\%
8 & 754\\%
9 & 802\\%
10 & 831\\%
\noalign{\smallskip}\hline%
\end{tabular}
\end{table}

The number of cases in which a feasible solution has been reached
slowly converges to 1000, monotonically increasing with each
additional loop. This indicates that the biased reconstruction in
the presented approach successfully adapts to events which are
difficult to assign to timeslots.

It should be noticed that the behavior of the approach for the other
benchmark instances is similar. This observation is however less
important, as a repetitive application of the simple constructive
approach will increase the percentage of cases in which a feasible
solution is reached, too. For instance {\tt comp05.ctt}, where not a
single feasible solution is found after the first loop, this does
not hold.

\section{\label{sec:iteration}Threshold Accepting based improvement}
\subsection{Description of the approach}
The constructive approach as described in
Section~\ref{sec:construction} only aims to identify a first
feasible assignment of events to timeslots, not taking into
consideration the resulting soft constraint violations. An iterative
procedure continues from here, searching for an optimal solution
with respect to the soft constraints.\\The formulation of the
approach is rather general. One of the reason for this is that while
we hope for a feasible assignment of all events, the constructive
approach does not guarantee it. Nevertheless, search for improved
solutions needs to continue at some point, and an approach that is
able to handle infeasible solutions is therefore required. Also, in
case of an infeasible first assignment, the procedure should be able
to later identify a feasible one.

In each step of the procedure, a number of randomly chosen events is
unassigned from the timetable and reinserted in the set
$\mathcal{E}^{u}$. A reassignment phase follows. Contrary to the
constructive approach, where events are selected based on whether
they are critical with respect to the available timeslots, events
are now randomly chosen from $\mathcal{E}^{u}$, each event with
identical probability. The choice of the timeslot follows the logic
as described in the constructive approach, prioritizing timeslots of
particular room classes. Again, we use the two possible preference
structures of rooms, $\mathcal{R}_{i1}$ over $\mathcal{R}_{i2}$ over
$\mathcal{R}_{i3}$, and $\mathcal{R}_{i2}$ over $\mathcal{R}_{i1}$
over $\mathcal{R}_{i3}$. Each of them is randomly chosen with
probability 0.5.

When evaluating timetables, two criteria are considered. First, the
number of unassigned timeslots (distance to feasibility) $hc$,
second, the total penalty with respect to the given soft constraints
$sc$. Comparison of solutions implies a lexicographic ordering of
the hard constraint violations $hc$ over the penalty function $sc$.
We therefore accept timetables minimizing the distance to
feasibility independent from the soft constraint count. This means
that in cases in which the initial construction phase is unable to
assign all events to timeslots, a later assignment of more events is
preferred independent from an increasing value of $sc$, closely
following the evaluation of solutions as given in the ITC 2007.\\In
case of identical distance to feasibility $hc$, inferior solutions
with respect to $sc$ are accepted up to a threshold. This idea has
been introduced by the Threshold Accepting metaheuristic
\cite{dueck:1990:article}, a simplified deterministic variant of
Simulated Annealing. Previous research has shown that
simplifications of Simulated Annealing may be very effective for
timetabling problems \cite{burke:2003:article}.

The implementation of the Threshold Accepting approach compares the
quality of neighboring solutions with the current best alternative,
permitting an acceptance of inferior alternatives up to the given
threshold. An alternative strategy would be the comparison with the
current solution instead of the globally best one. In this case
however, a subsequent acceptance of inferior solutions can happen,
and for that reason, the more restrictive acceptance strategy has
been chosen.

\subsection{First results and comparison with other approaches}
Different configurations of the algorithm have been tested on the
benchmark data from the ITC 2007. A first implementation has been
made available, however without optimizing the code with respect to
speed and efficiency. This has been done later, and the final
results for the ITC 2007, as reported later, are therefore
significantly better, simply because the final version of the
program allowed much faster computations. On an Intel Core 2 Quad
Q6600 2.4 GHz processor, equipped with 2 GB RAM, mounted on an ASUS
motherboard, 375 seconds of computing time have been allowed
for each test run.

Besides the determination of the number of reassigned events in each
iteration, which has been set to five, an appropriate choice of the
threshold needs to be made. Three different configurations of the
threshold are reported here, 0\% of $sc$, 1\%, and 2\%.

The following Table~\ref{tbl:results:improvement:1} gives the
obtained average values of the soft constraint penalties $sc$ for
three threshold configurations and compares the results to an
Iterated Local Search approach \cite{lourenco:2003:incollection}. In
this context, a threshold of 0\% leads to a hillclimbing algorithm
as only improving moves are accepted.\\The Iterated Local Search
approach consists of a hillclimbing algorithm (a Threshold Accepting
algorithm with threshold 0\%), perturbing the current solution after
a number of non-improving moves. Perturbations are done by a random
reassignment of five events. Contrary to the usual acceptance rule
with respect to the cost function $sc$, the perturbed alternative is
accepted in any case, and search continues from this new solution.
Two configurations of the Iterated Local Search Approach have been
implemented. The first variant, ILS 10k, starts pertubing after
10,000 non-improving moves, the other, ILS 3k, after 3,000
moves.

\begin{table}[!ht]
\caption{\label{tbl:results:improvement:1}Average values of $sc$}
\begin{tabular}{lrrrrr}
\hline\noalign{\smallskip}%
Instance & TA 0\% & TA 1\% & TA 2\%  & ILS 10k & ILS 3k\\%
\noalign{\smallskip}\hline\noalign{\smallskip}%
{\tt comp01.ctt} & {\bf 10} & 12 & 13 & 12 & 14  \\%
{\tt comp02.ctt} & 229 & {\bf 199} & 204 &218 &223 \\%
{\tt comp03.ctt} & 216 & {\bf 201} &213 &211 &202 \\%
{\tt comp04.ctt} & 134 & {\bf 126} &132 &138 &145 \\%
{\tt comp05.ctt} & 656 & {\bf 594} &657 &658 &641 \\%
{\tt comp06.ctt} & 199 & {\bf 177} &230 &196 &194 \\%
{\tt comp07.ctt} & {\bf 179} & 196 &316 &181 &185 \\%
\noalign{\smallskip}\hline%
\end{tabular}
\end{table}

On the basis of the obtained results, we conclude that a rather
small threshold of 1\% leads for most instances to the best average
results. There are some instances in which the Iterated Local Search
obtains good results, but TA 1\% is overall most promising.

It should be noticed that the choice of a percentage as a threshold
has been identified after experimenting with other algorithmic
variants. The main advantage of this approach appears to be that for
small values of $sc$ the algorithm behaves more like a hillclimbing
algorithm, while for larger values a larger threshold is derived.

\subsection{Results for the International Timetabling Competition ITC 2007}
The initial implementation of the algorithm has been optimized with
respect to execution speed, however keeping the methodological ideas
as described above. A significant improvement has been achieved, due
in particular to a delta-evaluation of the moves.

Table~\ref{tbl:results:ITC} gives the best results of the Threshold
Accepting algorithm with a threshold of 1\%. The results are based
on 30 trials with different random seeds. Each trial was allowed to
run for 375 seconds on the hardware mentioned above. The number of
evaluated solutions is given, too. In contrast the the initial
experiments, we now report results for 14 instances, seven of which
had been released a few weeks before the required submission of the
results.

\begin{table}[!ht]
\caption{\label{tbl:results:ITC}Best results and the used seeds (out
of 30 trials)}
\begin{tabular}{lcrrr}
\hline\noalign{\smallskip}%
Instance & seed & hard constraint & soft constraint & evaluations\\%
 & & violations & violations\\
\noalign{\smallskip}\hline\noalign{\smallskip}%
{\tt comp01.ctt} & 130 & 0 &   5 & 13,072,619\\
{\tt comp02.ctt} & 112 & 0 & 108 &  8,547,980\\
{\tt comp03.ctt} & 119 & 0 & 115 &  9,211,859\\
{\tt comp04.ctt} & 128 & 0 &  67 & 10,352,548\\
{\tt comp05.ctt} & 119 & 0 & 408 &  6,512,059\\
{\tt comp06.ctt} & 117 & 0 &  94 &  8,631,146\\
{\tt comp07.ctt} & 113 & 0 &  56 &  7,673,851\\
{\tt comp08.ctt} & 129 & 0 &  75 &  9,881,464\\
{\tt comp09.ctt} & 119 & 0 & 153 &  9,248,758\\
{\tt comp10.ctt} & 122 & 0 &  66 &  8,386,538\\
{\tt comp11.ctt} & 111 & 0 &   0 & 13,468,229\\
{\tt comp12.ctt} & 103 & 0 & 430 &  6,782,742\\
{\tt comp13.ctt} & 104 & 0 & 101 &  9,838,210\\
{\tt comp14.ctt} & 122 & 0 &  88 &  9,693,538\\
\noalign{\smallskip}\hline%
\end{tabular}
\end{table}

It can be seen that the approach leads to reasonable results, and
that the best results of the improved code are significantly better
than the ones of the first implementation. For some instances, {\tt
comp01.ctt} and {\tt comp11.ctt}, particularly good solutions are
found. Others such as {\tt comp05.ctt} and {\tt comp12.ctt} have
best found alternatives with soft constraint penalties that are
still quite large. Based on the observed improvement in comparison
to the first implementation, we can conclude that efficiency of the
implementation plays an important role for the final results.

The following Table~\ref{tbl:ITC:all:results} gives the average
results of the top five competitors of ITC 2007, track 3. The
columns are sorted in descending order of the overall ranking, thus
showing the results of Thomas M\"{u}ller in the leftmost column. In
brief, our approach ranked 4th overall. When closer analyzing the
obtained results, it becomes clear that the approaches of the first
three finalists did indeed lead to comparable superior results. In
relation to the approach of Clark, Henz, and Love, our
implementation of the Threshold Accepting algorithm turned out to be
better, however not for all test instances.\\Unfortunately, we do
not have any information about the algorithms of the other
finalists. Consequently, the possibilities of drawing precise
conclusions are limited. Nevertheless, we suspect that the top three
ranked programs are substantially better than our Threshold
Accepting implementation, simply because the average results are
superior. This raises the question whether the observed differences
are due to a better (faster) implementation, or due to better
algorithmic ideas. Longer optimization runs are therefore carried
out in the following, allowing a better convergence of the
metaheuristic without the immediate pressure of terminating the
search after only 375 seconds.

\begin{table}
\caption{\label{tbl:ITC:all:results}Average results of the top five
competitors of ITC 2007, track 3}
\begin{tabular}{lrrrrr}
\hline\noalign{\smallskip}%
Instance & M\"{u}ller & Lu, Hao & Atsuta, & Geiger & Clark,\\
& (USA) & (France) & Nonobe, & (Germany) & Henz, Love\\
& & & Ibaraki & & (Singapore)\\
& & & (Japan)\\
\noalign{\smallskip}\hline\noalign{\smallskip}%
Rank: & 1 & 2 & 3 & 4 & 5\\
\noalign{\smallskip}\hline\noalign{\smallskip}%
{\tt comp01.ctt} &   5.0 &   5.0 &   5.1 &   6.7 &  27.0\\%
{\tt comp02.ctt} &  61.3 &  61.2 &  65.6 & 142.7 & 131.1\\%
{\tt comp03.ctt} &  94.8 &  84.5 &  89.1 & 160.3 & 138.4\\%
{\tt comp04.ctt} &  42.8 &  46.9 &  39.2 &  82.0 &  90.2\\%
{\tt comp05.ctt} & 343.5 & 326.0 & 334.5 & 525.4 & 811.5\\%
{\tt comp06.ctt} &  56.8 &  69.4 &  74.1 & 110.8 & 149.3\\%
{\tt comp07.ctt} &  33.9 &  41.5 &  49.8 &  76.6 & 153.4\\%
{\tt comp08.ctt} &  46.5 &  52.6 &  46.0 &  81.7 &  96.5\\%
{\tt comp09.ctt} & 113.1 & 116.5 & 113.3 & 164.1 & 148.9\\%
{\tt comp10.ctt} &  21.3 &  34.8 &  36.9 &  81.3 & 101.3\\%
{\tt comp11.ctt} &   0.0 &   0.0 &   0.0 &   0.3 &   5.7\\%
{\tt comp12.ctt} & 351.6 & 360.1 & 361.6 & 485.1 & 445.3\\%
{\tt comp13.ctt} &  73.9 &  79.2 &  76.1 & 110.4 & 122.9\\%
{\tt comp14.ctt} &  61.8 &  65.9 &  62.3 &  99.0 & 105.9\\%
{\tt comp15.ctt} &  94.8 &  84.5 &  89.1 & 160.3 & 138.0\\%
{\tt comp16.ctt} &  41.2 &  49.1 &  50.2 &  92.6 & 107.3\\%
{\tt comp17.ctt} &  86.6 & 100.7 & 107.3 & 143.4 & 166.6\\%
{\tt comp18.ctt} &  91.7 &  80.7 &  73.3 & 129.4 & 126.8\\%
{\tt comp19.ctt} &  68.8 &  69.5 &  79.6 & 132.8 & 125.4\\%
{\tt comp20.ctt} &  34.3 &  60.9 &  65.0 &  97.5 & 179.3\\%
{\tt comp21.ctt} & 108.0 & 124.7 & 138.1 & 185.3 & 185.8\\%
\noalign{\smallskip}\hline%
\end{tabular}
\end{table}

\subsection{Convergence in longer runs}
In contrast to the optimization runs for the ITC 2007, we allow in
the following experiments the evaluation of 100 million timetables
before terminating the algorithm. Again, 30 trials have been carried
out, and Table~\ref{tbl:results:longruns} gives the best found
solutions out of all test runs.

\begin{table}[!ht]
\caption{\label{tbl:results:longruns}Best results after 100,000,000
evaluations (out of 30 trials)}
\begin{tabular}{lrr}
\hline\noalign{\smallskip}%
Instance & hard constraint & soft constraint\\%
 & violations & violations\\
\noalign{\smallskip}\hline\noalign{\smallskip}%
{\tt comp01.ctt} & 0 &   5\\
{\tt comp02.ctt} & 0 &  91\\
{\tt comp03.ctt} & 0 & 108\\
{\tt comp04.ctt} & 0 &  53\\
{\tt comp05.ctt} & 0 & 359\\
{\tt comp06.ctt} & 0 &  79\\
{\tt comp07.ctt} & 0 &  36\\
{\tt comp08.ctt} & 0 &  63\\
{\tt comp09.ctt} & 0 & 128\\
{\tt comp10.ctt} & 0 &  49\\
{\tt comp11.ctt} & 0 &   0\\
{\tt comp12.ctt} & 0 & 389\\
{\tt comp13.ctt} & 0 &  91\\
{\tt comp14.ctt} & 0 &  81\\
\noalign{\smallskip}\hline%
\end{tabular}
\end{table}

Obviously, the Threshold Accepting algorithm did not converge after
only 375 seconds. Rather big improvements can be seen for most
instances, sometimes improving the best solution by 25\% ({\tt
comp10.ctt}). For the instances with large values of $sc$, {\tt
comp05.ctt} and {\tt comp12.ctt}, improvements are possible, but the
absolute values remain rather high. We suspect that
these instances possess properties that complicate the
identification of timetables with small soft constraint violations.
Recalling that instance {\tt comp05.ctt} was problematic with
respect to the identification of a feasible assignment in the
initial experiments, this is however not surprising.\\No
improvements are possible for instance {\tt comp01.ctt}, and of
course for instance {\tt comp11.ctt}.

In comparison to the three top ranked finalists of ITC 2007,
inferior overall results are found, even when allowing the execution
of 100,000,000 evaluations. Independent from the personal
programming skills of the competitors, which are unknown to us and
difficult to assess, we suspect that the performance of the
approaches is mainly due to the algorithms as such.

\section{\label{sec:conclusions}Summary and conclusions}
The article presented an approach for curriculum-based course
timetabling, employing the general idea of the Threshold Accepting
metaheuristic. The methodological concepts are rather
problem-independent as only simple removals and reassignments of
events from and to the timetable are carried out during search.

Initial experiments with a first implementation indicated that small
values of the threshold present a good parameter setting. Comparison
studies with a simple hillclimbing algorithm and an Iterated Local
Search Algorithm have been carried out. In brief, the Threshold
Accepting variant with a threshold of 1\% appeared to be most
promising.

Comparisons of the short runs for the International Timetabling
Competition 2007 with long runs reveal that the proposed algorithm
does not converge within the given time limit. More time for
computations is needed, and further improvements of the concept are
certainly possible.

We are confident that a fair contribution to the ITC 2007 has been
made. In comparison to the other participants of the ITC 2007, our
approach ranked 4th overall. However, a considerable gap to the
average results of the top three contributions became obvious, and
we are looking forward to read the articles describing these
approaches. Nevertheless, good solutions are found, in some cases
even in short time. We find optimal solutions for instance {\tt
comp11.ctt}, and a very good one for instance {\tt comp01.ctt}.


\begin{acknowledgements}
The author would like to thank three anonymous referees for their
helpful comments.
\end{acknowledgements}

\bibliographystyle{plain}
\bibliography{lit_bank,lit_bank_nv,lit_bank_tt}

\end{document}